%% file: egpaper_final.tex
\documentclass[10pt,twocolumn,letterpaper]{article}

\input{newcommands2}

\usepackage{cvpr}
\usepackage{times}
\usepackage{epsfig}
\usepackage{graphicx}
\usepackage{amsmath}
\usepackage{amssymb}

\usepackage{subcaption}
\usepackage{booktabs}
\usepackage{arydshln}

\usepackage[normalem]{ulem}

\usepackage[pagebackref=true,breaklinks=true,letterpaper=true,colorlinks,bookmarks=false]{hyperref}

\cvprfinalcopy % *** Uncomment this line for the final submission

\def\cvprPaperID{5757} % *** Enter the CVPR Paper ID here

\let\oldtwocolumn\twocolumn
\renewcommand\twocolumn[1][]{%
    \oldtwocolumn[{#1}{
    \begin{center}
          \vspace{-.3in}
           \includegraphics[width=\textwidth]{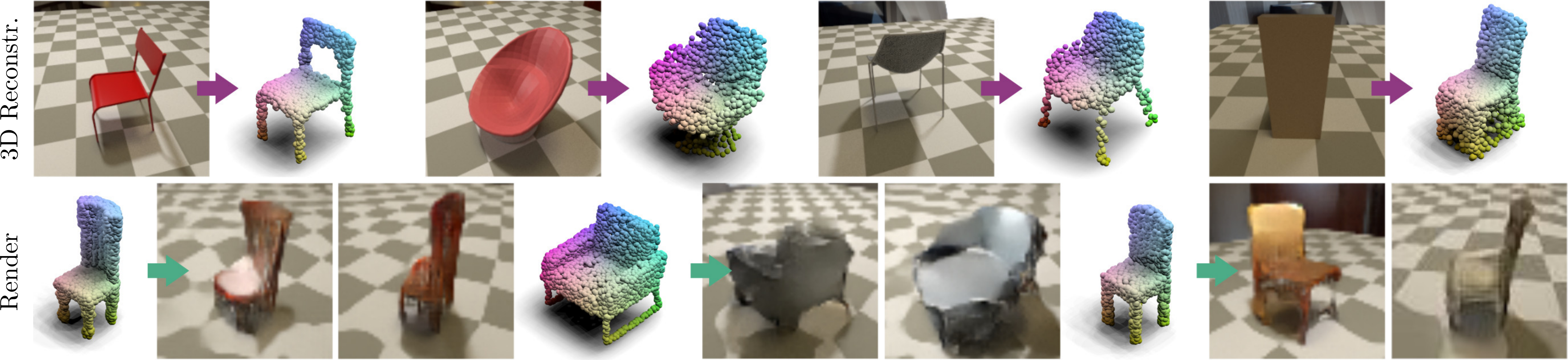}
           \vspace{-0.25in}
           \captionof{figure}{\small{We propose \mbox{C-Flow},  a conditioning scheme for flow-based generative models applicable to many different domains. The figure shows the results of modeling the conditional distributions {\em image $\leftrightarrow$ 3D point cloud}. In the top row we apply this model for 3D reconstruction ({\em image $\rightarrow$ point cloud}), and in the bottom row for rendering new images ({\em point cloud $\rightarrow$ image}). Our model allows sampling multiple times from this conditional distribution to generate several renderings of the same point cloud. }}
           
           \label{fig:teaser}
        \end{center}
    }]
}

\begin{document}

%%%%%%%%% TITLE
\title{C-Flow: Conditional Generative Flow Models for Images and 3D Point Clouds \vspace{-.8em}}
\author{Albert Pumarola$^{1,}$\thanks{Work done while interning at Google.}\hspace{0.8cm}
Stefan Popov$^{2}$ \hspace{0.8cm}
Francesc  Moreno-Noguer$^{1}$ \hspace{0.8cm}
Vittorio Ferrari$^{2}$ \hspace{0.8cm}\\
$^{1}$Institut de Rob\`otica i Inform\`atica Industrial, CSIC-UPC, Barcelona, Spain\\
$^{2}$Google Research, Z\"{u}rich, Switzerland
}

%\maketitle
%\thispagestyle{empty}
\vspace{-1em}
\maketitle

\begin{abstract}
Flow-based generative models have highly desirable  properties like exact log-likelihood evaluation and exact latent-variable inference, however they are still in their infancy and have not received as much attention as alternative generative models.
In this paper, we introduce \mbox{C-Flow}, a novel conditioning scheme that brings normalizing flows to an entirely new scenario with great possibilities for multi-modal data modeling.
\mbox{C-Flow} is based on a parallel sequence of invertible mappings in which a
source flow  guides the target flow  at every step, enabling fine-grained control over the generation process.
We also devise a new strategy to model unordered 3D point clouds that,  in combination with the conditioning scheme, makes it possible to address 3D reconstruction from a single image and its inverse problem of rendering an image given a point cloud.
We demonstrate our conditioning method to be very adaptable, being also applicable to image manipulation, style transfer and multi-modal image-to-image mapping in a diversity of domains, including RGB images, segmentation maps and edge masks.

\end{abstract}

\input{01_intro}

\input{02_related}

\input{03_method}

\input{04_experiments}

\input{05_conclusions}

{\small
\bibliographystyle{ieee_fullname}
\bibliography{egbib}
}

\end{document}

%% file: newcommands2.tex
\newcommand{\comment}[1]{}

\newcommand{\bzero}{\textbf{0}}

\newcommand{\bff}{\mathbf{f}}
\newcommand{\bg}{\mathbf{g}}
\newcommand{\bh}{\mathbf{h}}

\newcommand{\bs}{\mathbf{s}}

\newcommand{\bt}{\mathbf{t}}

\newcommand{\bx}{\mathbf{x}}

\newcommand{\bz}{\mathbf{z}}

\newcommand{\bI}{\mathbf{I}}

\newcommand{\bN}{\mathbf{N}}

\newcommand{\btheta}{\boldsymbol{\theta}}
\newcommand{\bvartheta}{\boldsymbol{\vartheta}}

\newcommand{\bphi}{\boldsymbol{\phi}}
\newcommand{\bvarphi}{\boldsymbol{\varphi}}

%% file: 01_intro.tex
\vspace{-1em}
\section{Introduction}
\vspace{-.5em}
Generative models have become extremely popular in the machine learning and computer vision communities. Two main actors currently prevail in this scenario, Variational Autoencoders (VAEs)~\cite{KingmaW13} and especially Generative Adversarial Networks (GANs)~\cite{goodfellow2014generative}. 
In this paper we focus on a different family, the so-called flow-based generative models~\cite{dinh2014nice}, which remain under the shadow of VAEs and GANs despite offering very appealing properties. Compared to other generative method, flow-based models build upon a sequence of reversible mappings between the input and latent space that  allow for
(1)  exact latent-variable inference and log-likelihoood evaluation,
(2) efficient and parallelizable inference and synthesis and
(3) useful and simple data manipulation by operating directly on the latent space.

The main contribution  of this paper is a novel approach to condition normalizing flows, making it possible to perform multi-modality transfer tasks which have so far not been explored under the umbrella of flow-based generative models.  For this purpose, we introduce \mbox{C-Flow}, a framework consisting of two parallel flow branches, interconnected across their reversible functions using conditional coupling layers and trained with an invertible cycle consistency. This scheme allows guiding a source domain towards a target domain guaranteeing the satisfaction of the aforementioned properties of flow-based models.
Conditional inference is then implemented in a simple manner, by (exactly) embedding the source sample into its latent space, sampling a point from a Gaussian prior, and then propagating them through the learned normalizing flow.
For example, for the application of synthesizing multiple plausible photos given a semantic segmentation mask, each image  is generated by jointly propagating the segmentation embedding and a random point drawn from a prior distribution across the learned flow.

Our second contribution is a strategy to enable flow-based methods to model unordered 3D point clouds. Specifically, we introduce
(1) a re-ordering of the 3D data points according to a Hilbert sorting scheme,
(2) a global feature operation compatible with the reversible scheme,
and (3) an invertible cycle consistency that penalizes the Chamfer distance.
Combining this strategy with the proposed conditional scheme we can then 
address tasks such as shape interpolation, 3D object reconstruction from an image, and rendering an image given a 3D point cloud (Fig.~\ref{fig:teaser}).

Importantly, our new conditioning scheme enables a wide range of tasks beyond 3D point cloud modeling. In particular, we are the first flow-based model to show mapping between a large diversity of domains, including image-to-image, pointcloud-to-image, edges-to-image segmentation-to-image and their inverse mappings.  Also, we are the first to demonstrate application in image content manipulation and style transfer tasks.

We believe our conditioning scheme, and its ability to deal with a variety  of domains, opens the door to building general-purpose and easy to train solutions. We hope all this will spur future research in the domain of flow-based generative models.

%% file: 02_related.tex
\vspace{-1mm}
\section{Related Work}

\vspace{-1.5mm}
\paragraph{Flow-Based Generative Models.}
Variational Auto-Encoders (VAEs)~\cite{KingmaW13} and Generative Adversarial Networks (GANs)~\cite{goodfellow2014generative} are the most studied deep generative models so far.
VAEs  use deep  networks as  function approximators and maximize a lower bound on the data log-likelihood to model a continuous latent variable with intractable posterior distribution~\cite{razavi2019generating, sohn2015learning, Lassner_2017_ICCV}.
GANs, on the other hand, circumvent the need for dealing with likelihood estimation by leveraging an adversarial strategy. %
While GANs' versatility has made possible  advances in many applications~\cite{isola2017image, pumarola2019ganimation, zhu2017unpaired, corona2020ganhand, caelles2019fast, agustsson2018generative}, their training is
unstable~\cite{radford2015unsupervised} and requires careful hyper-parameter tuning.

Flow-based generative models~\cite{dinh2014nice, rezende2015variational} have received little attention compared to GANs and VAEs, despite offering very attractive properties such as the ability to estimate  exact log-likelihood,  efficient synthesis and exact latent-variable inference. Further advances have been proposed in RealNVP~\cite{dinh2016density} by introducing the affine coupling layers and in Glow~\cite{kingma2018glow}, through an   architecture with 1x1 invertible convolutions for image generation and editing. These works have been later applied to audio generation~\cite{prenger2019waveglow, kim2018flowavenet, yamaguchi2019adaflow, Serra19ARXIV},  image modeling~\cite{haoliang2019dual, Grover2019AlignFlowCC, chen2019beautyglow} and video prediction~\cite{kumar2019videoflow}. 

Some recent works have proposed  strategies for conditioning normalizing flows by combining them with other generative models. For instance,~\cite{liu2019conditional, Grover2019AlignFlowCC}  combine flows with GANs. These models, however, are more 
difficult to train as adversarial losses tend to introduce instabilities. 
Similarly, for the specific application of video prediction, ~\cite{kumar2019videoflow} enforces an autoregressive model onto the past latent variables to predict them in the future. Dual-Glow~\cite{haoliang2019dual} uses a  conditioning scheme  for MRI-to-PET brain scan mapping by concatenating the prior distribution of the source image with the latent variables of the target image.

In this paper, we introduce  a novel  mechanism to condition flow-based generative models by enforcing a source-to-target coupling at every transformation step instead of only feeding the source information into the target prior distribution.
As we show experimentally, this enables fine-grained control over the modeling process (Sec.~\ref{sec:experimentation}).

\vspace{1mm}
\noindent{\bf Modeling and reconstruction of 3D Shapes.}
The success of deep learning has spurred a large number of discriminative approaches for 3D reconstruction~\cite{choy20163d, Pumarola_2018_CVPR, wang2018pixel2mesh, sinha2017surfnet, groueix2018atlasnet, yang2018foldingnet}. These techniques, however, only learn direct mappings between  output shapes and   input images. Generative models, in contrast, capture the actual shape distribution from the training set,   enabling not only  to reconstruct new test images, but also to  sample new shapes from the learned distribution. 
There exist several works along this line. For instance, GANs have been used in Wu~\etal~\cite{wu2016learning} to model objects in a voxel representation; Hamu~\etal~\cite{ben2018multi} used them  to model body parts;
and Pumarola~\etal~\cite{pumarola20193dpeople} to learn the manifold of geometry images representing clothed 3D bodies.
Auto-encoders~\cite{dai2017shape, stutz2018learning} and VAEs~\cite{fan2017point, bagautdinov2018modeling, mescheder2019occupancy, henderson19ijcv} have also been applied to model 3D data.
More recently, Joon Park~\etal~\cite{park2019deepsdf} used auto-decoders~\cite{bojanowski2017optimizing, fan2018matrix} to represent shapes with continuous volumetric fields. All previous techniques are not bijective, and thus,  not directly applicable to our model.

PointFlow~\cite{yang2019pointflow} is the only approach that uses normalizing flows to model 3D data. They learn a generative model for point clouds 
by first modeling the distribution of object shapes and then applying normalizing flows to model the point cloud distribution for each shape.  This strategy, however, cannot condition the shape, preventing PointFlow from  being used in applications such as 3D reconstruction and rendering. Also, its inference time is very high, as point clouds are generated one point at a time,  while we generate the entire point cloud in one forward pass.

%% file: 03_method.tex
\vspace{-1mm}
\section{Flow-Based Generative Model}
\vspace{-1.9mm}
Flow-based generative models aim to approximate an unknown true data distribution $\bx \sim p^*(\bx)$ from a limited set of observations $\{\bx^{(i)}\}_{i=1}^N$.
The data is modeled by learning an invertible transformation $\bg_{\btheta}(\cdot)$ mapping a latent space with tractable density $p_{\bvartheta}(\bz)$ to $\bx$:
\begin{equation}
\label{eq:nf_sampling}
    \bz \sim p_{\bvartheta}(\bz), \quad \bx = \bg_{\btheta}(\bz),
\end{equation}
where $\bz$ is a latent variable and $p_{\bvartheta}(\bz)$ is typically a  Gaussian distribution $\bN(\bz; 0,\bI)$. The function $\bg_{\btheta}$, commonly known as a normalizing flow~\cite{rezende2015variational}, is bijective, meaning that given a data point $\bx$ its latent-variable $\bz$ is computed as:
\begin{equation}
\label{eq:nf_bijective}
    \bz = \bg^{-1}_{\btheta}(\bx),
\end{equation}
where $\bg^{-1}_{\btheta}$ is composed of a sequence of $K$ invertible transformations $\bg^{-1} = \bg_1^{-1} \circ \bg_2^{-1} \circ \cdots \circ \bg_{K}^{-1}$ defining a mapping between $\bx$ and $\bz$ such that:
{
\setlength{\belowdisplayskip}{5pt} \setlength{\belowdisplayshortskip}{5pt}
\setlength{\abovedisplayskip}{5pt} \setlength{\abovedisplayshortskip}{5pt}
\begin{equation}
\label{eq:inv_transforms}
\bx \triangleq \bh_0 \overset{\bg_1^{-1}}{\longleftrightarrow} \bh_1 \overset{\bg_2^{-1}}{\longleftrightarrow} \bh_2 \cdots \overset{\bg_K^{-1}}{\longleftrightarrow} \bh_K \triangleq \bz,
\end{equation}
}
%$K$ being a fixed hyper-parameter.
The goal of generative models is to find the parameters $\btheta$ such that $p_{\btheta}(\bx)$ best approximates $p^*(\bx)$.
Explicitly modeling such probability density function is usually intractable, but using the normalizing flow mapping of Eq.~\eqref{eq:nf_sampling} under the change of variable theorem,
we can compute the exact log-likelihood for a given data point $\bx$ as:
{
\setlength{\belowdisplayskip}{5pt} \setlength{\belowdisplayshortskip}{5pt}
\setlength{\abovedisplayskip}{5pt} \setlength{\abovedisplayshortskip}{5pt}
\begin{align}
    \log p_{\btheta}(\bx) &= \log p_{\bvartheta}(\bz) +  \log|\det(\partial\bz/\partial\bx)|  \\
    &= \log p_{\bvartheta}(\bz) + \sum_{i=1}^K \log|\det(\partial\bh_i/\partial\bh_{i-1})| \label{eq:log_like}
\end{align}
}
where $\partial\bh_i/\partial\bh_{i-1}$ is the Jacobian matrix of $\bg^{-1}_i$ at $\bh_{i-1}$ and the Jacobian determinant measures the change of log-density made by $\bg^{-1}_i$ when transforming $\bh_{i-1}$ to $\bh_{i}$.  Since we can now compute the exact log-likelihood, the training criterion of flow-based generative model is
directly
the negative log-likelihood over the observations.
Note that optimizing over the actual log-likelihood of the observations is  more stable and informative than doing it over a lower-bound of the log-likelihood for VAEs, or minimizing the adversarial loss in GANs. This is one of the major virtues of flow-based approaches.

\vspace{-1mm}
\section{Conditional Flow-Based Generative Model}
\vspace{-1mm}
\begin{figure}[t!]
\centering
\includegraphics[width=\linewidth]{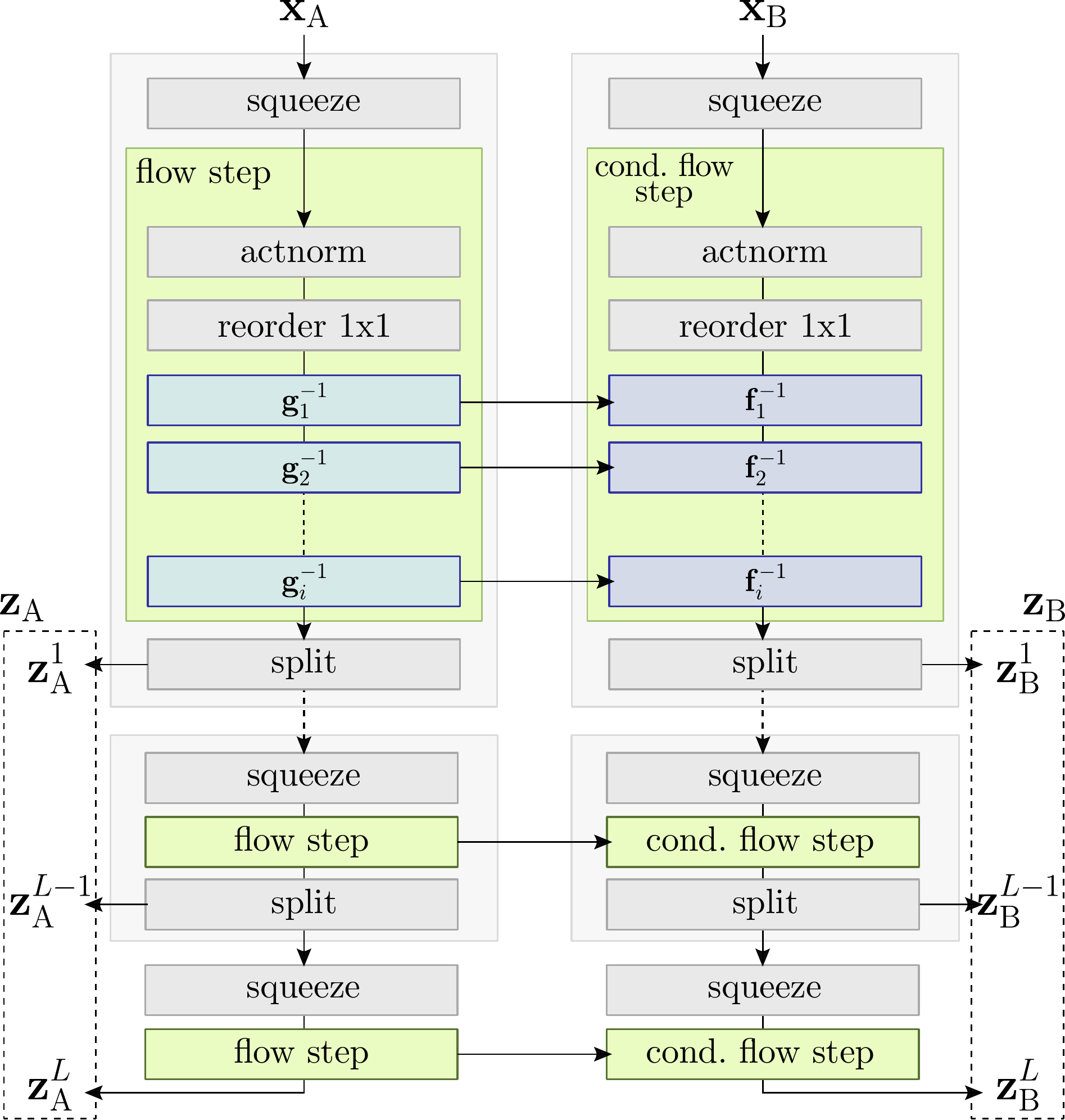}\\
\vspace{-.4em}
\caption{\small{{\bf The C-Flow model}  consists of two parallel flow branches mutually interconnected with conditional coupling layers. This scheme allows sampling $\bx_\text{B}$ conditioned on $\bx_\text{A}$. For a detailed description on functions in grey refer to~\cite{kingma2018glow}.} %
}
\vspace{-1.4em}
\label{fig:model}
\end{figure}

\begin{figure*}
    \vspace{-1em}
    \centering
    \begin{subfigure}[b]{0.38\linewidth}        %
        \centering
        \includegraphics[width=\linewidth]{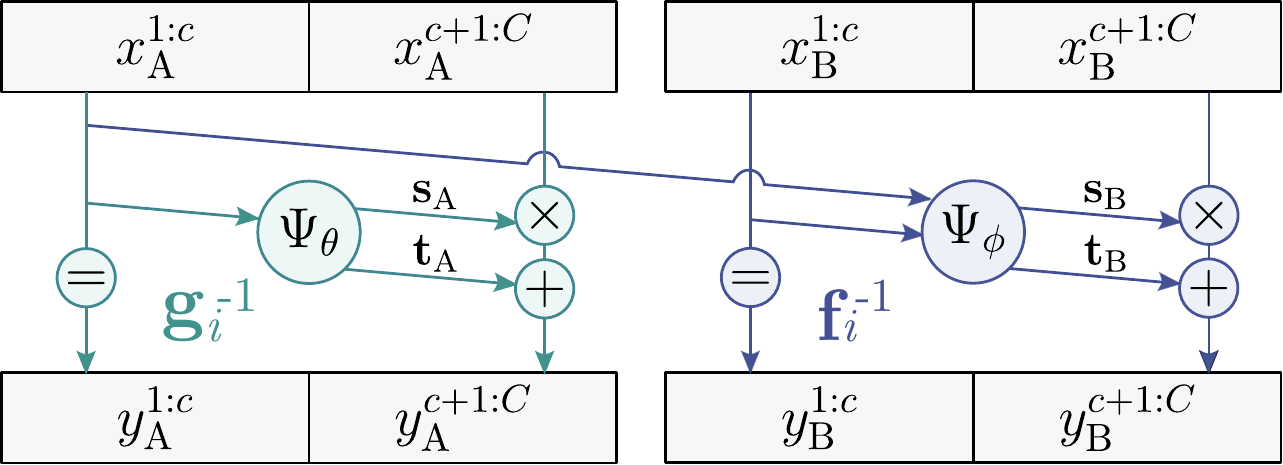}
        \caption{\small{Forward Propagation}}
    \end{subfigure}
    \hspace{16mm}
    \begin{subfigure}[b]{0.38\linewidth}        %
        \centering
        \includegraphics[width=\linewidth]{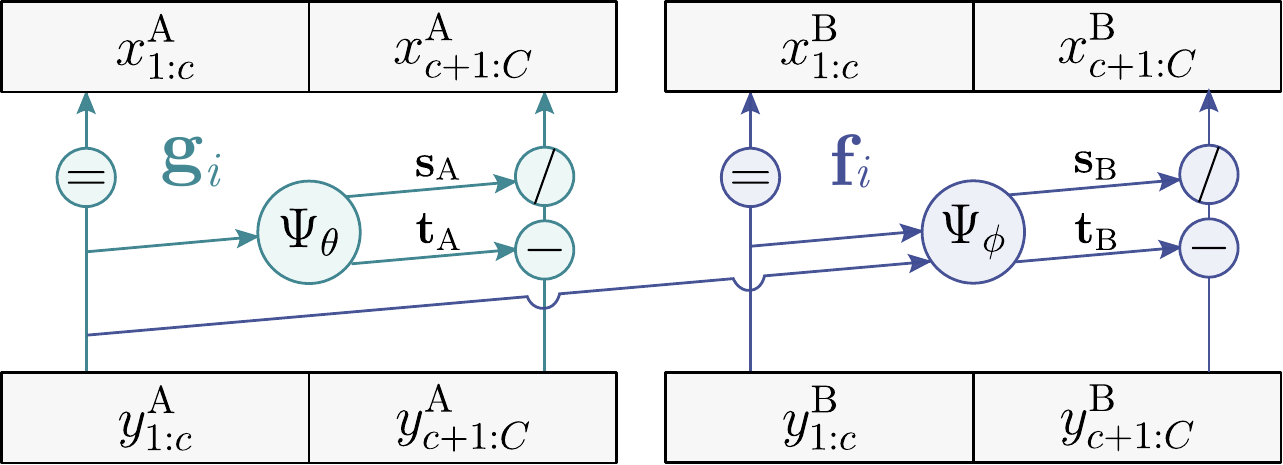}
        \caption{\small{Backward Propagation}}
    \end{subfigure}
    \vspace{-1.7mm}
    \caption{\small{{\bf Conditional coupling layer for forward and backward propagation.} Given two input tensors $\bx_\text{A}$ and $\bx_\text{B}$, the proposed conditional coupling layer transforms
      the second half of $\bx_\text{B}$   conditioned on the first halves of $\bx_\text{A}$ and $\bx_\text{B}$. The first halves of all tensors are not updated. By sequentially concatenating these bijective operations we can transform data points $x$ into their latent representation $y$ (forward propagation) and vice versa (backward propagation).}
    }
    \label{fig:cond_coupling}
    \vspace{-1em}
\end{figure*}

Given a true data distribution $(\bx_\text{A}, \bx_\text{B}) \sim p^*(\bx_\text{A}, \bx_\text{B})$.
Our goal is to learn a model for  $\bx_\text{B} \sim p^*(\bx_\text{B}|\bx_\text{A})$ to map sample points from domain $A$ to domain $B$.
For example, for the application of 3D reconstruction, $\bx_\text{A}$ would be an image and $\bx_\text{B}$ a 3D point cloud.
To this end, we propose a conditional flow-based generative model extending~\cite{dinh2016density, kingma2018glow}. Our L-levels model, learns both distributions with two bijective transformations $\bg_{\btheta}$ and $\bff_{\bphi}$ (Fig.~\ref{fig:model}):
\begin{align}
    \bz_\text{A} \sim p_{\bvartheta}(\bz_\text{A}), \quad \bz_\text{B} &\sim p_{\bvarphi}(\bz_\text{B}) \\
    \bx_\text{A} = \bg_{\btheta}(\bz_\text{A}), \quad \bx_\text{B} &= \bff_{\bphi}(\bz_\text{B}| \bz_\text{A}) \\
    \bz_\text{A} = \bg^{-1}_{\btheta}(\bx_\text{A}), \quad \bz_\text{B} &= \bff^{-1}_{\bphi}(\bx_\text{B}| \bx_\text{A})
\end{align}
where $\bz_\text{A}$ and $\bz_\text{B}$ are latent-variables, and $p_{\bvartheta}(\bz_\text{A})$ and $p_{\bvarphi}(\bz_\text{B})$ are tractable spherical multivariate Gaussian distributions with
learnable mean and variance. Note that conditioning on $\bz_A$ or $\bx_A$ is equivalent, as they are related by a bijective transformation.

We then define the mapping $\mathcal{M}$ to sample $\bx_\text{B}$ conditioned on $\bx_\text{A}$,
as a three-step operation:
\begin{align}
    \bz_\text{A} &= \bg^{-1}_{\btheta}(\bx_\text{A}) && \text{encode condition } \bx_\text{A} \\
    \quad \bz_\text{B} &\sim p_{\bvarphi}(\bz_\text{B}) && \text{sample latent-variable } \bz_\text{B} \\
    \bx_\text{B} &= \bff_{\bphi}(\bz_\text{B}| \bz_\text{A})  && \text{generate $\bx_\text{B}$ cond. on $\bx_\text{A}$}
\end{align}

In the following subsections we describe how this conditional framework  is implemented. Sec.~\ref{sec:cond_coupling}  discusses the  foundations of the conditional coupling layer  we propose to map source to target data using invertible functions, and how its Jacobian is computed. Sec.~\ref{sec:cond_nn}  describes the architecture we define for the practical implementation of the coupling layers. Sec.~\ref{sec:inv_cycle} presents an invertible cycle consistency loss   introduced to further stabilize the training process.
Finally, in Sec.~\ref{sec:total_loss} we define the total training loss.

\subsection{Conditional Coupling Layer}
\label{sec:cond_coupling}
\vspace{-1mm}

When designing the conditional coupling layer we need to fulfill the constraint that each transformation has to be bijective and tractable. As shown in~\cite{dinh2014nice, dinh2016density}, both these issues can be overcome by  choosing  transformations  with triangular Jacobian. In this case their determinant is calculated as the product of  diagonal terms, making the computation tractable and ensuring invertibility. Motivated by these works, we propose an extension of their coupling layer to account for cross-domain conditioning. A schematic of the proposed layers is shown in Fig.~\ref{fig:cond_coupling}. Formally, let us   define $y \triangleq h_i$ and $x \triangleq h_{i-1}$. We then write the invertible function $\bff^{-1}$ to transform  a data point $\bx_\text{B}$ based on $\bx_\text{A}$ as follows:
\begin{align}
\begin{cases}
   y^{1:c}_\text{B} \quad\: = x^{1:c}_\text{B} & \\
   y^{c+1:C}_\text{B} = x^{c+1:C}_\text{B} \odot \exp\left ( s \left(x^{1:c}_\text{A}, x^{1:c}_\text{B} \right) \right) + t\left ( x^{1:c}_\text{A}, x^{1:c}_\text{B} \right ) & ,
   \nonumber
\end{cases}
\end{align}
where $C$ is the number of channel dimensions in both data points, $\odot$ denotes element-wise multiplication and $s$ and $t$ are the scale and translation functions \mbox{$(\mathbb{R}^c, \mathbb{R}^c) \mapsto \mathbb{R}^{C-c}$}. We set $c=C/2$ in all experiments. For $\bff$ it is not strictly necessary to split $\bx_\text{A}$ to ensure bijectiveness. However, by doing  so  we  highly  reduce  the  computational  requirements.

The inverse $\bff$ of the   conditional coupling layer is:
{
\setlength{\belowdisplayskip}{0pt} \setlength{\belowdisplayshortskip}{0pt}
\begin{align}
\begin{cases}
   x^{1:c}_\text{B} \quad\: = y^{1:c}_\text{B} \\
   x^{c+1:C}_\text{B} = \big(y^{c+1:C}_\text{B} - t\left ( y^{1:c}_\text{A}, y^{1:c}_\text{B}\right)\big)  \odot \exp\left ( s \left(x^{1:c}_\text{A}, x^{1:c}_\text{B} \right) \right), %
   \label{eq:backward}
\end{cases}
\end{align}
}
and its Jacobian:
\begin{equation}
\frac{\partial y_\text{B}}{\partial x^\top} = \left[\begin{array}{cc}
\bI_{c} & \bzero \\
\frac{\partial y^{c+1:C}_\text{B}}{\partial \left(x^{1:c}\right)^\top} & \text{diag}\big(\exp\left(s\left(x^{1:c}_\text{A}, x^{1:c}_\text{B} \right)\right)\big) \\
\end{array} \right] \nonumber
,
\end{equation}
where $\bI_{c} \in \mathbb{R}^{c \times c}$ is an identy matrix.
Since the Jacobian is a triangular matrix, its determinant can be calculated efficiently
as the product of the diagonal elements.
Note that it is not required to compute the Jacobian of the functions $s$ and $t$, enabling them to be arbitrarily complex. In practice, we implement these functions using a  convolutional neural network $\Psi(\cdot)$ that returns both  $\log (\bs)$ and $\bt$. %

\subsection{Coupling Network Architecture}
\label{sec:cond_nn}
\vspace{-1mm}
We next describe the architecture of $\Psi_{\btheta}(\cdot)$ and $\Psi_{\bphi}(\cdot)$ used to regress the affine transform applied at every  conditional coupling layer at each $\bg_i$ and $\bff_i$ respectively.
We build upon the stack of three 2D convolution layers proposed by~\cite{kingma2018glow}.
The first two layers have a filter size of $3\times3$ and $1\times1$ with $512$ output channels followed by actnorm~\cite{kingma2018glow} and a ReLU activation.
The third layer regresses the final scale and translation by applying a 2D convolutional layer with filter size $3\times3$ initialized with zeros
such that each affine transformation at the beginning of training is equivalent to an identity function. %

For the transformation $\bg^{-1}_i(x_\text{A})$ we exactly use this architecture, but for $\bff^{-1}_i(x_\text{B} | x_\text{A})$ we extend it to take into account the conditioning $x_\text{A}$.
Concretely, in $\bff^{-1}_i$, $x_\text{B}$ is initially transformed by two convolution layers, like the first two of $\bg^{-1}_i$. Then, $x_\text{A}$ is adapted with a channel-wise affine transform implemented by a $1\times1$ convolution. Finally, its output is added to the transformed $x_\text{B}$. To ensure a similar contribution of  $x_\text{A}$ and  $x_\text{B}$ %
their activations are normalized with actnorm so that they operate in the same range. A final $3\times3$ convolution regresses the conditional coupling layer operators $\log(\bs_\text{B})$ and $\bt_\text{B}$.

\subsection{Invertible Cycle Consistency}
\label{sec:inv_cycle}
\vspace{-1.5mm}

We train our model to maximize the log-likelihood of the training dataset. However, likewise in  GANs learning~\cite{Pathak_2016_CVPR, pix2pix2016}, we found beneficial to add a loss encouraging the generated and real samples to be similar in L1.
To do so, we exploit the fact that our model is made of bijective transformations, and introduce what we call an invertible cycle consistency. This operation can be summarized as follows:
\begin{equation}
     \{\bx_\text{A}, \bx_\text{B}\} \overset{\bg^{-1}, \bff^{-1}}{\longrightarrow} \{\bz_\text{A}, \bz_\text{B}\} \rightarrow \{\bz_\text{A}, \hat{\bz}_\text{B}\} \overset{\bff}{\longrightarrow} \hat{\bx}_\text{B}.
\end{equation}
Concretely, the data points observations ($\bx_\text{A}$, $\bx_\text{B}$) are initially mapped into their latent variables ($\bz_\text{A}$, $\bz_\text{B}$), where each  variable is composed of an $L$-level stack. As demonstrated in~\cite{dinh2016density} the first levels encode the high frequencies (details) in the data, and the last levels the low frequencies.

We then resample the first $L-1$ dimensions of $\bz_\text{B}$ from a Gaussian distribution, \ie $\bz_\text{B}= [\bz_1, \dots, \bz_L] \rightarrow \hat{\bz}_\text{B}= [\bN(0,\bI)_1, \dots, \bN(0,\bI)_{L-1}, \bz_L]$. By doing this, $\hat{\bz}_\text{B}$ is only retaining the lowest frequencies of the original $\bz_\text{B}$.

As a final  step, we invert $\bff^{-1}$, to recover  $\hat{\bx}_\text{B} = \bff(\hat{\bz}_\text{B} | {\bz}_\text{A})$ and penalize its L1 difference w.r.t the original ${\bx}_\text{B}$. What we are essentially doing is to force the model to use information from the condition ${\bx}_\text{A}$
so that the recover sample $\hat{\bx}_\text{B}$ is as similar as possible to the original ${\bx}_\text{B}$.
Note that if reconstructed $\hat{\bz}_\text{B}$ based on the entire latent variable, the recovered sample  would be identical to the original ${\bx}_\text{B}$ because $\bff$ is bijective, and this loss would be meaningless.

\vspace{-1mm}
\subsection{Total Loss}
\label{sec:total_loss}
\vspace{-1.5mm}

Formally, denoting the training pairs of observations as $\{\bx^{(i)}_\text{A},
\bx^{(i)}_\text{B}\}_{i=1}^N$, the model parameters are learned by minimizing
the following loss function:%
\begin{equation}
\label{eq:final_loss}
\frac{1}{N}\sum_{i=1}^N \left[
  -\log p_{\btheta,\bphi}(\bx^{(i)}_\text{A}, \bx^{(i)}_\text{B})
  + \lambda \left \| \bx^{(i)}_\text{B} - \hat{\bx}^{(i)}_\text{B} \right \|_1
\right]
\end{equation}

The first term maximizes the joint likelihood of the data observations. With our
design, it also maximizes the conditional likelihood of
$\bx_\text{B}|\bx_\text{A}$ and thus forces the model to learn the desired
mapping. To show this, we apply the law of total probability and we factor it into:
\begin{equation}
-\sum^N_{i=1}{\log p_{\btheta}(\bx^{(i)}_\text{A})} - \sum^N_{i=1}{\log p_{\bphi}(\bx^{(i)}_\text{B} | \bx^{(i)}_\text{A})}
\end{equation}
Due to the diagonal structure of the Jacobians, the marginal likelihood of
$\bx_\text{A}$ depends only on $\btheta$ (first sum), while the conditional of
$\bx_\text{B}|\bx_\text{A}$, only on $\bphi$. Maximizing the joint
likelihood thus maximizes both likelihoods independently.

The second term in \eqref{eq:final_loss} minimizes the cycle consistency loss.
$\lambda$ is a hyper-parameter balancing the terms. This loss is fully
differentiable, and we provide details on how we optimize it in
Sec.~\ref{sec:implementation}.

\begin{figure}[t!]
\vspace{-1.5em}
\centering
\includegraphics[width=\linewidth]{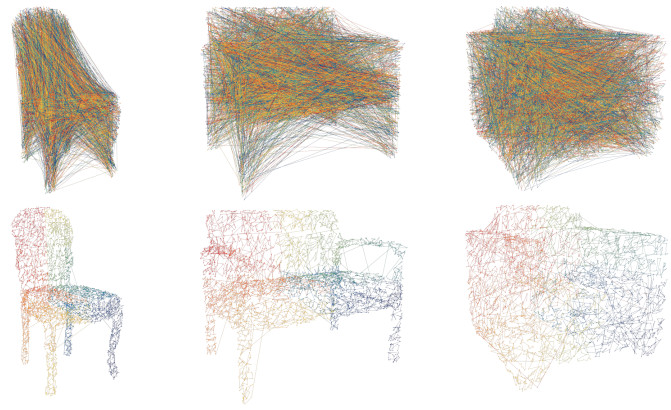}\\
\vspace{-.9em}
\caption{\small{{\bf Sorting 3D point clouds.} Point clouds corresponding to three different chairs. The colored line connects all points based on their ordering. {\bf Top:} Unordered. {\bf Bottom:} Applying the proposed
sorting strategy.
Note how the coloring is consistent across samples even for point clouds with different topology.}}
\label{fig:hilbert}
\vspace{-1.5em}
\end{figure}

\section{Modeling Unordered 3D Point Clouds} \label{sec:modelingpcl}
\vspace{-1.5mm}

The model described so far  can handle input data represented on regular grids but it fails to model unordered 3D point clouds, whose lack of spatial neighborhood ordering prevents convolutions from being applied.
To process point clouds with deep networks, a common practice is to apply \textit{symmetry operations}~\cite{qi2017pointnet} that create fixed-size tensors of global features describing the entire point cloud sample. These operations  require extracting point-independent features followed by a max-pool, which is not invertible and  not applicable to normalizing flows.  Another alternative would be the graph convolutional networks~\cite{wu2019comprehensive}, although their high computational cost makes them not suitable for our scheme of multiple coupling layers.
We propose a three-step mechanism to enable modeling 3D point clouds:

\vspace{1mm}
\noindent{\bf (i) Approximate Sorting with Space-Filling Curves.} \mbox{C-Flow} is based on convolutional layers which require input data with a local neigboorhood consistent across samples. To fulfill this condition on unordered point clouds, we propose to sort them   based on proximity.  As discussed in~\cite{qi2017pointnet}, for high dimensional spaces it is not possible to produce a perfect ordering stable to point perturbations. In this paper we therefore consider using the approximation provided by the  Hilbert's space-filling curve algorithm~\cite{Hilbert1935}. For each training sample, we project its points into a 3D Hilbert curve and reorder them based on their ordering along the curve (Fig.~\ref{fig:hilbert}).
Notice that not only we can establish a neighborhood relationship but also a semantically-stable ordering (\eg in Fig.~\ref{fig:hilbert} the chair's right-leg is always blue). To the best of our knowledge there is no previous work using such preprocessing for point clouds.

\begin{figure}[t!]
\centering
\includegraphics[width=0.7\linewidth]{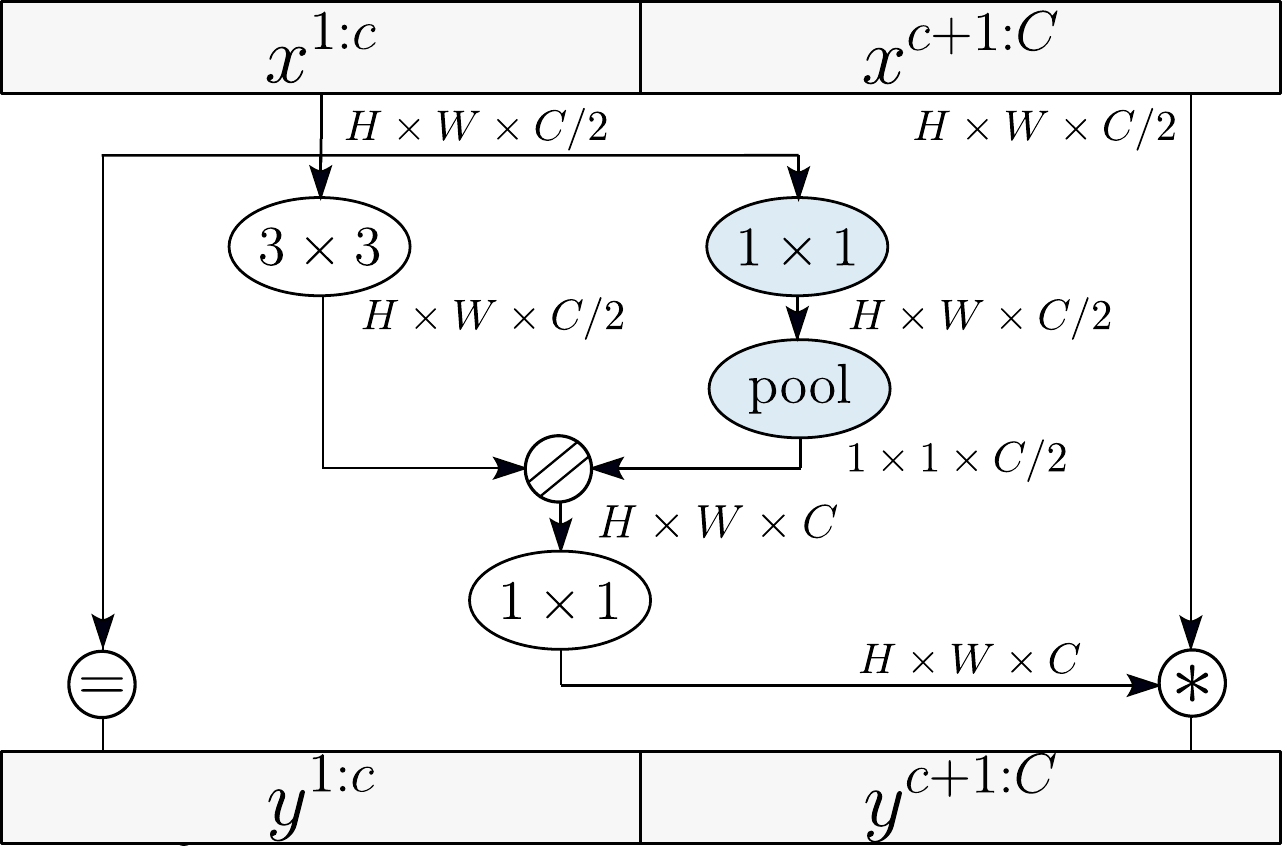}\\
\caption{\small{{\bf Approximating global features in point clouds.} When dealing with point clouds (reordered and reshaped to a $H\times W\times 3$ size and using $c = C/2$) we approximate, with operations in blue, global features in coupling layers while still being invertible. $\circledast$ stands for affine transformation where the first $C/2$ input channels are the scale and the other half the translation.
}}
\label{fig:global}
\vspace{-1.4em}
\end{figure}

\vspace{1mm}
\noindent{\bf (ii) Approximating Global Features.}
Hilbert Sort is not sufficient to model 3D data because of a major issue: it splits the space into equally sized quadrants and the Hilbert curve will cover all points in a quadrant before moving to the next. As a consequence, two points that were originally close in space, but lie near the boundaries of two different quadrants, will  end up far away in the final ordering.
To mitigate this effect we  extend the proposed coupling network architecture (Sec.~\ref{sec:cond_nn}) with an approximate but invertible version of the global features proposed in~\cite{qi2017pointnet} that describe the whole point cloud.
Concretely, we first resample and reshape the reordered point cloud to form $H\times W\times 3$ matrices (in practice we use the same size as that of the images). Then we  approximate the global descriptors of~\cite{qi2017pointnet} through a  $1 \times 1$ convolution to extract point-independent features followed by a max-pool applied only over the first half of the point cloud features
$x^{1:c}$ (Fig.~\ref{fig:global}).
The coupling layer remains bijective because during the backward propagation the approximated global features can be recovered using a similar strategy as in Eq.~\eqref{eq:backward}.

\vspace{1mm}
\noindent{\bf (iii) Symmetric Chamfer Distance for Cycle Consistency.}
For the specific case of point clouds, we observed that when penalizing the invertible cycle consistency with L1 the model converged to a mean Hilbert curve. Therefore, for point clouds, we  substitute L1  by the symmetric Chamfer distance , which computes the mean Euclidean distance between the ground-truth point cloud $\bx_\text{B}$ and the recovered $\hat{\bx}_\text{B}$.
\begin{figure}[t!]
\centering
\vspace{-2em}
\includegraphics[width=\linewidth]{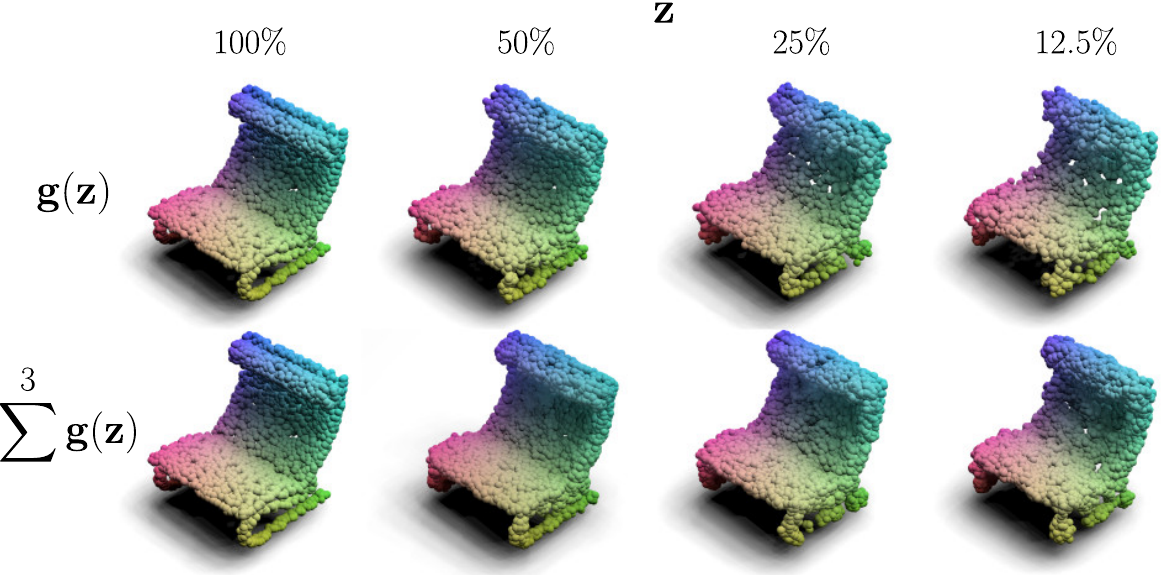}\\
\vspace{-.7em}
\caption{\small{{\bf Embedding 3D points clouds.} {\bf Top:} Reconstruction with partial embeddings. {\bf Bottom:} Reconstruction with
three iterations of backward propagations of partial embeddings.}}
\label{fig:repr}
\vspace{-1.2em}
\end{figure}

\vspace{-1mm}
\section{Implementation Details}
\label{sec:implementation}
\vspace{-1.5mm}

Due to memory restrictions, we train with image samples of $64 \times 64$  resolution. For 3D point clouds, to maintain the same architecture as in images, we reshape each point cloud sample (list of $64^2$ points) to $64 \times 64$. At test time we also regress $64^2$ 3D points per forward pass. Our implementation builds upon that of Glow~\cite{kingma2018glow}. We use Adam with learning rate $1e^{-6}$, $\beta_1=0.85$, $\beta_2=0.007$ and batch size $4$.  The multi-scale architecture consists of $L=4$ levels with 12 flow steps per level ($K=4*12$ in Eq.~\eqref{eq:inv_transforms})
each and $2 \times$ squeezing operations. For conditional sampling we found additive coupling ($s(\cdot)=1$) to be more stable during training than affine transformation.
The prior distributions $p_{\bvartheta}(\bz_\text{A})$ and $p_{\bvarphi}(\bz_\text{B})$ are initialized with mean 0 and variance 1. The rest of weights are randomly initilized from a normal distribution with mean $0$ and std $0.05$.
$\lambda=10$ in Eq.~\eqref{eq:final_loss}.
As in previous likelihood-based generative models~\cite{parmar2018image, kingma2018glow}, we observed that sampling from a reduced-temperature prior improves the  results. To do so, we multiply the variance of $p_{\bvarphi}(\bz_\text{B})$ by $T=0.9$. The model is trained with 4 GPUs P-100 for 10 days.

%% file: 04_experiments.tex
\vspace{-2mm}
\section{Experimental Evaluation}
\label{sec:experimentation}
\vspace{-1.5mm}

We next  evaluate our system on diverse tasks: (1)  Modeling point clouds (Sec.~\ref{sec:points_emb}), (2)  3D reconstruction and rendering  (Sec.~\ref{sec:embedding}), (3) Image-to-image mapping  in a variety of domains and datasets (Sec.~\ref{sec:img2img}), and (4) Image manipulation and style transfer  (Sec.~\ref{sec:imgtrans}).

\vspace{-1mm}
\subsection{Modeling 3D Point Clouds}
\label{sec:points_emb} 
\vspace{-1mm}

We  evaluate the potential of our approach to model 3D point clouds on ShapeNet~\cite{chang2015shapenet}. For this task, we do not consider the full conditioning scheme and  only use one of the branches of \mbox{C-Flow} in Fig.~\ref{fig:model}, which we denote as \mbox{C-Flow*}.

In our first experiment we study the representation capacity of unknown shapes, formally  defined as the ability to retain the information after mapping forward and backward between the original  and latent spaces. For this purpose,  we first map  a real point cloud $\bx$ to the latent space $\bz = \bg^{-1}_{\btheta}(\bx)$. The full-size embedding $\bz=[\bz_1,\ldots,\bz_L]$ has as many dimensions as the input ($HWC$). Then we progressively remove information from $\bz$  by  replacing their left-most $l$  components with samples drawn from a Gaussian distribution, \ie $\hat\bz=[\bN(0,\bI)_1, \dots, \bN(0,\bI)_{l},\bz_{l+1}, \dots, \bz_L]$.  Note that the embedding size $L-l$ can be set at test time with no need to retrain, making tasks like point cloud compression straightforward. Finally we   map back this  embedding to the original point cloud space $\hat\bx = \bg_{\btheta}(\bz)$ and compare to $\bx$.

\begin{figure}[t!]
\centering
\vspace{-2em}
\includegraphics[width=\linewidth]{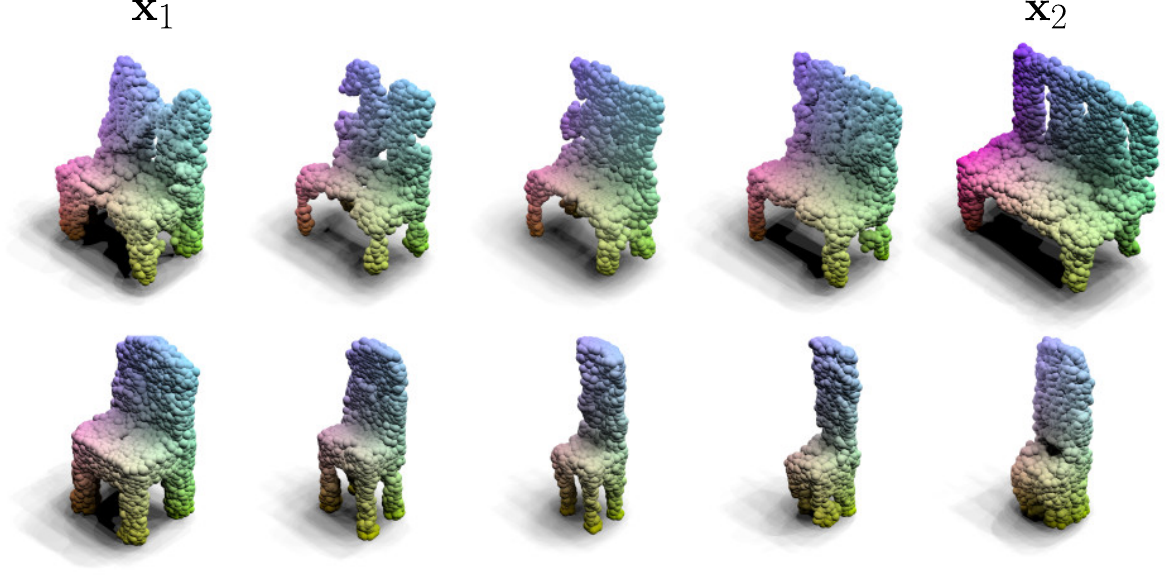}\\
\vspace{-.7em}
\caption{\small{{\bf Interpolation.} Results of interpolating two 3D point clouds $\bx_1$ and $\bx_2$ in the learned latent space.}} 
\label{fig:interp}
\vspace{-.5em}
\end{figure}

\input{table_1.tex}

Tab.~\ref{table:quant} reports the Chamfer Distance (CD)   for different embedding sizes. CD is computed by densely resampling the input mesh with up to $10^7$ vertices. The plain version of \mbox{C-Flow*} (no conditioning, no sorting, no global features) is equivalent to Glow~\cite{kingma2018glow}.  This   version is consistently improved when introducing the   sorting and global features strategies (Sec.~\ref{sec:modelingpcl}). The error decreases gracefully as we increase the embedding size, and importantly, when using  the full size embedding we obtain a perfect recovering (Fig.~\ref{fig:repr}-top). This is a virtue of the bijective models, and is not a trivial property. 
Tab.~\ref{table:quant}  also reports the numbers of %
AtlasNet~\cite{groueix2018atlasnet} and DeepSDF~\cite{park2019deepsdf}, showing that our approach achieves competitive results. This comparison is only indicative as the representation used are inherently different (~\cite{groueix2018atlasnet} parametric and~\cite{park2019deepsdf} continuous surface).

Recall that the left-most components 
%randomly sampled 
in $\bz$ encode the shape high details. We exploit this property to generate point clouds with an arbitrarily large number of points by performing multiple backward propagations ($\hat\bx = \bg_\theta(\hat{\bz})$) of a partial embedding $\hat{\bz}$ (Fig.~\ref{fig:repr}-bottom). Every time we propagate, we recover a new set of  3D points allowing to progressively improve the density of the reconstruction. 

Another task that can be addressed with C-Flow is shape interpolation in the latent space (Figure~\ref{fig:interp}).

\begin{figure*}[t!]
\vspace{-1.5em}
\centering
\includegraphics[width=\linewidth]{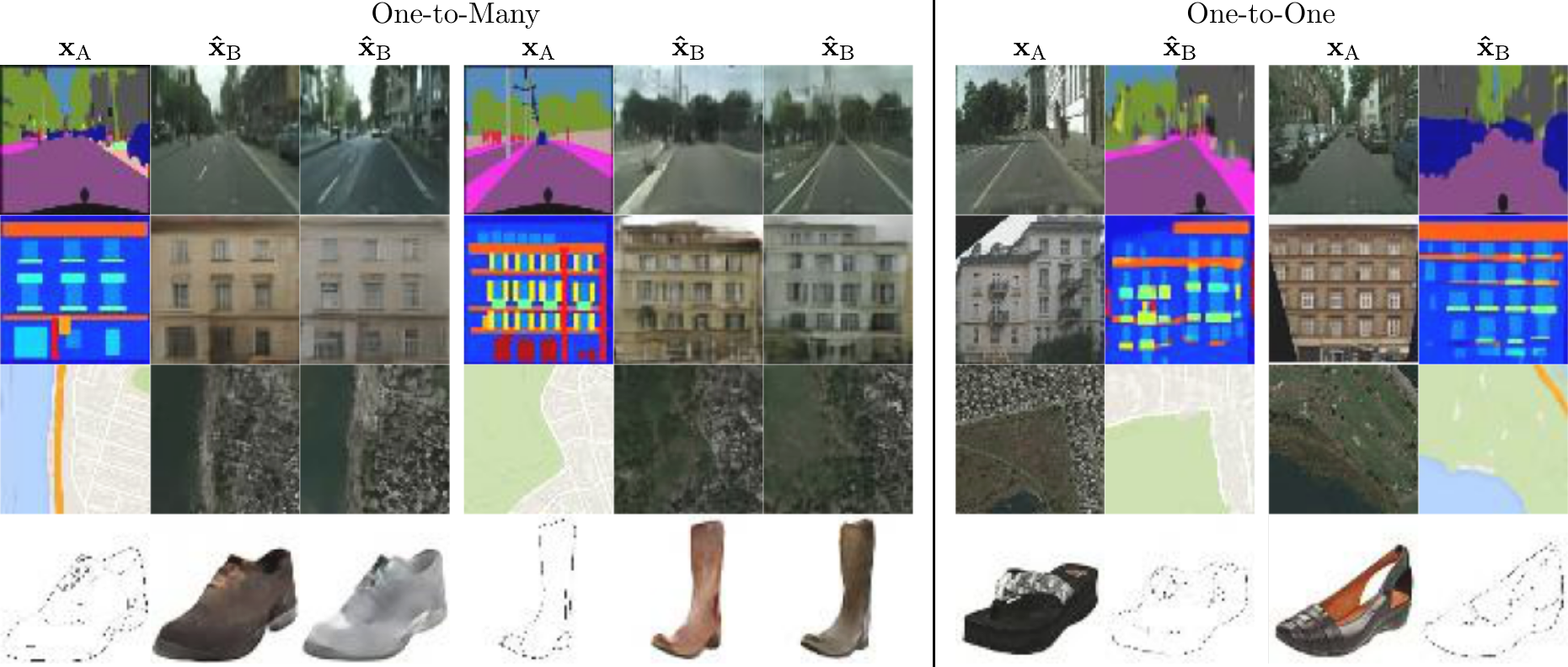}\\
\vspace{-.3em}
\caption{\small{{\bf Image-to-Image.} Results from  $64 \times 64$ image-to-image mappings on a variety of domains.
$\bx_\text{A}$: source image; $\mathbf{\hat{x}}_\text{B}$: generated image in the target domain. The examples on the left correspond to target domains with high variability that when sampled multiple times generate different images. In the examples on the right the target domain has a small variability and the sampling becomes deterministic.
}}
\label{fig:img2img}
\vspace{-1.3em}
\end{figure*}

\input{table_2.tex}

\subsection{3D Reconstruction \& rendering}
\label{sec:embedding}

We next evaluate the ability of \mbox{C-Flow} to model the conditional distributions
(1) {\em  image $\rightarrow$  point cloud} 
, which enables to perform 3D reconstruction from a single image; and
(2) {\em  point cloud  $\rightarrow$  image}, which is its inverse problem of rendering an image given a 3D point cloud.
Fig.~\ref{fig:teaser} shows qualitative results on the {\em Chair} class of ShapeNet. In the top row   our model is able to generate plausible 3D reconstructions of unknown objects even under strong self-occlusions (top-right example).
The second row depicts results for rendering, which highlights another advantage of our model: it allows  sampling multiple times from the conditional
distribution to produce several images of the same object which exhibit different properties (\eg viewpoint or texture).

In Table~\ref{table:3drec} we  compare  C-Flow with other single-image 3D reconstruction methods 3D-R2N2~\cite{choy20163d}, PSGN~\cite{fan2017point}, Pix2Mesh~\cite{wang2018pixel2mesh}, AtlasNet~\cite{groueix2018atlasnet} and ONet~\cite{mescheder2019occupancy}.
We evaluate 3D reconstruction in terms of the Chamfer distance (CD) with the ground truth shapes.
Our approach 
(last row) performs on par with~\cite{choy20163d,fan2017point,wang2018pixel2mesh} and it is slightly below the state-of-the-art techniques specifically designed for 3D reconstruction~\cite{groueix2018atlasnet,mescheder2019occupancy}.

\input{table_3.tex}

\begin{figure*}[t!]
 \vspace{-1.5em}
 \centering
    \begin{subfigure}[b]{\linewidth}      
        \centering
        \includegraphics[height=3.4cm, width=\linewidth]{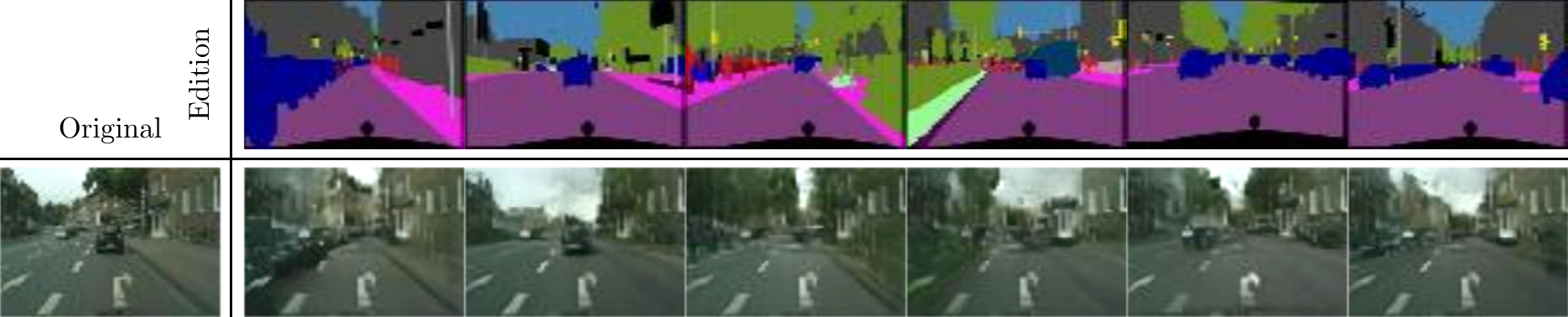}  
        \caption{\small{Image content manipulation}}
    \end{subfigure}
    \begin{subfigure}[b]{\linewidth}
        \centering
        \includegraphics[height=3.2cm, width=\linewidth]{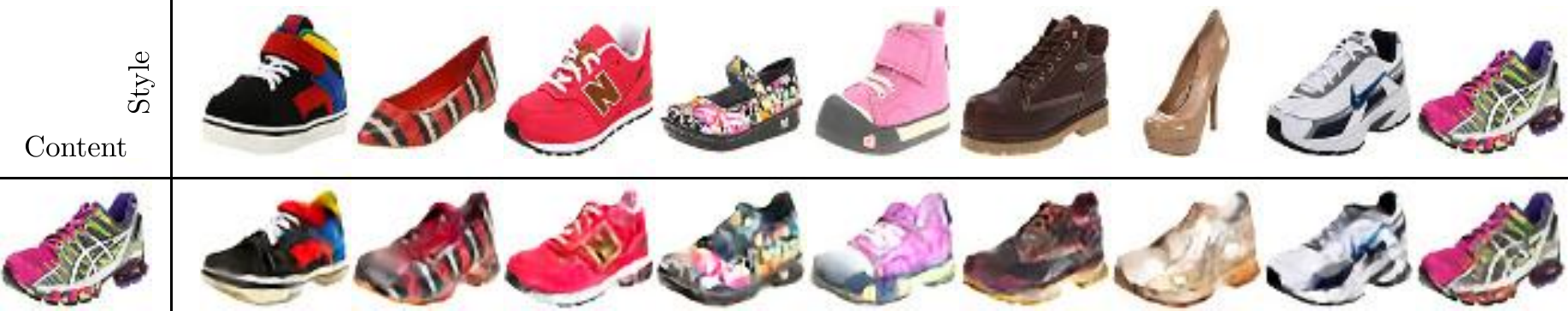}  
        \caption{\small{Style transfer}}
    \end{subfigure}
    \vspace{-2em}
    \caption{\small{{\bf Other applications.} Sample results on $64 \times 64$ image manipulation and style transfer. The model was not retrained for these tasks, and we used the same training weights to perform image-to-image in Fig.~\ref{fig:img2img}.}}
    \label{fig:translation}
    \vspace{-1.2em}

\end{figure*}

With the same model, we can also render images from point clouds. To the best of our knowledge, no previous work can perform such mapping.
While a few approaches do render point clouds~\cite{meshry2019neural, Aliev2019NeuralPG, pittaluga2019revealing}, they hold on strong assumptions of knowing the RGB color per point and the camera calibration  to project the point cloud onto the image plane. Table~\ref{table:3drec} also reports an ablation study about the different operations we devised to handle 3D point clouds, namely  sorting the point cloud (Sort), approximating global features (GF-Coupling) and inverse cycle consistency with chamfer distance (CD). In this case, evaluation is reported using Inception Score (IS)~\cite{salimans2016improved} and Bits Per Dimension (BPD) which is equivalent to the negative log2-likelihood typically used to report flow-based methods performance.
Results show a performance boost when using each of these components, and especially when combining them.

\subsection{Image-to-Image mappings}
\label{sec:img2img}
\vspace{-2mm}
 We  evaluate the ability of C-Flow to perform multi-domain image-to-image mapping:
{\em segmentation $\leftrightarrow$ street views} trained on Cityscapes~\cite{cordts2016cityscapes}, {\em structure $\leftrightarrow$ facade} trained on CMP Facades~\cite{Tylecek13}, {\em map $\leftrightarrow$ aerial photo} trained on~\cite{pix2pix2016} and {\em edges $\leftrightarrow$ shoes} trained on~\cite{yu2014fine, zhu2016generative, pix2pix2016}.
Fig.~\ref{fig:img2img}-left shows mappings in which the target domain has a wide variance and  multiple sampling generates different results (\eg a semantic segmentation map can map to several grayscale images).
Fig.~\ref{fig:img2img}-right examples have a target domain with a narrower variance, and despite multiple samplings the generated images are very similar (\eg given an  image its segmentation is well defined).  %

Table~\ref{table:im2im}  reports quantitative evaluations using Structural Similarity (SSIM)~\cite{wang2004image}, and again BPD and IS. 
When introducing the invertible cycle consistency loss (Sec.~\ref{sec:inv_cycle}) the model does not improve its compression abilities (BPD) but  improves in terms of structural similarity (SSIM) and semantic content (IS). %
It is worth mentioning that while GANs have shown impressive image-to-image mapping results, even at high resolution~\cite{wang2018high}, ours is the first work 
that can address such tasks using normalizing flows.

\subsection{Other Applications} 
\label{sec:imgtrans}
\vspace{-2mm}
Finally, we demonstrate the versatility of \mbox{C-Flow} 
being the first flow-based method capable of performing style transfer and image content manipulation (Fig.~\ref{fig:translation}). 
Importantly, the model was {\em not retrained} for these specific tasks, and we use the same parameters learned to perform image-to-image mappings (Sec.~\ref{sec:img2img}).
For image manipulation we use the weights of {\em segmentation $\rightarrow$ street view} and for style transfer  those of {\em edges $\leftrightarrow$ shoes}.
Formally, let the domain $A$ to be the structure (\eg segmentation mask) and the domain $B$  to be the image (\eg street view). Then, image manipulation is achieved via three operations:
\begin{align}
    \bz_\text{B}^1 &= \bff^{-1}_{\bphi}(\bx_\text{B}^1| \bx_\text{A}^1) && \text{encode original image } \bx_\text{B}^1 \\
    \bz_\text{A}^2 &= \bg^{-1}_{\btheta}(\bx_\text{A}^2) && \text{encode desired structure } \bx_\text{A}^2 \\
    \bx_\text{B}^2 &= \bff_{\bphi}(\bz_\text{B}^1| \bz_\text{A}^2)  && \text{synthesise new image $\bx_\text{B}^2$} 
\end{align}

Note that 
%following this generation approach 
we are no longer conditioning 
based only on $A$, as in Sec.~\ref{sec:img2img}, 
%and 
now the synthesised image is jointly conditioned on $A$ (for structure) and $B$ (for texture). 

To perform style transfer, we first transform the content image into its structure $\bx_\text{A}^2$. For instance, in Fig.~\ref{fig:translation}-bottom, the content of the {\em shoe} is initially mapped onto its {\em edge} structure with the {\em shoes $\rightarrow$ edges} weights. Then, we apply the same procedure as we did for image manipulation using the {\em edges $\rightarrow$ shoes} weights, 
setting $\bx_\text{A}^1$ to be the structure of the content image and $\bx_\text{B}^1$ the style image.

%% file: table_1.tex
\begin{table}[t!]
\vspace{-.5em}
\setlength{\tabcolsep}{4pt} % general space between cols (6pt standard)
\centering
\resizebox{0.4\textwidth}{!}{%
%\sisetup{detect-weight=true}
\begin{tabular}{lcccc}
\toprule
% Method & $100\%$ & $50\%$ & $25\%$ & $12.5\%$ \\
Method & $100\%$ & $50\%$ & $25\%$ & $12.5\%$ \\
\cmidrule(lr){1-1}  \cmidrule(lr){2-5}
% \cmidrule(lr){2-5}
    C-Flow* $\equiv$ Glow~\cite{kingma2018glow}  & 0.00  & 0.39 & 0.39 & 0.39 \\
    C-Flow* + Sort & 0.00 & 0.19 & 0.21 & {\bf 0.22} \\
    C-Flow* + Sort + GF-Coupling & {\bf 0.00} & {\bf 0.14} & {\bf 0.18} & 0.31 \\
\hdashline
    AtlasNet-Sph.~\cite{groueix2018atlasnet} &  \multicolumn{4}{c}{0.75}  \\
    AtlasNet-25~\cite{groueix2018atlasnet}  &  \multicolumn{4}{c}{0.37} \\
    DeepSDF~\cite{park2019deepsdf} &  \multicolumn{4}{c}{0.20} \\
\bottomrule
\end{tabular}
}
\vspace{-.4em}
\caption{\small{{\bf Representing 3D point clouds}. Chamfer distance when recovering point clouds with partial embeddings. For all C-Flow* we change the embedding size at test,  with no further training. The percentages are with respect to the input  dimension (4096).  For AtlasNet and DeepSDF  we provide the results from~\cite{park2019deepsdf}. }} 
\label{table:quant}
\vspace{-1.5em}
\end{table}

%% file: table_2.tex
\begin{table}[t!]
% \vspace{-.2em}
\setlength{\tabcolsep}{4pt} % general space between cols (6pt standard)
\centering
\resizebox{0.4\textwidth}{!}{%
%\sisetup{detect-weight=true}
\begin{tabular}{lc|cc}
\toprule
& Image $\rightarrow$ PC & \multicolumn{2}{c}{Image $\leftarrow$ PC} \\
Method & CD$\downarrow$  & BPD$\downarrow$ & IS$\uparrow$ \\
\cmidrule(lr){1-1}  \cmidrule(lr){2-2} \cmidrule(lr){3-4}
    3D-R2N2~\cite{choy20163d} & 0.27 & - & -  \\
    PSGN~\cite{fan2017point} & 0.26 & - & -  \\
    Pix2Mesh~\cite{wang2018pixel2mesh} & 0.27 & - & -  \\
    AtlasNet~\cite{groueix2018atlasnet} & {\bf 0.21} & - & -  \\
    ONet~\cite{mescheder2019occupancy} & 0.23 & - & -  \\
\cmidrule(lr){1-1}  \cmidrule(lr){2-2} \cmidrule(lr){3-4}
    C-Flow  & 0.86 &  4.38 &  1.80  \\
    C-Flow + Sort & 0.52 & {\bf 2.77}  &  2.41 \\
    C-Flow + Sort + GF-Coupling & 0.49 & 2.87 &   {\bf 2.61}  \\
    C-Flow + Sort + GF-Coupling + CD & {\bf 0.26} & - & -  \\
\bottomrule
\end{tabular}
}
\caption{\small{{\bf 3D Reconstruction and rendering.} $\downarrow$: the lower the better, $\uparrow$: the higher the better. C-Flow is the first approach able to   render images from point clouds. The same model can be used to perform 3D reconstruction from images. The results of all other methods are obtained from their original papers.}}
\vspace{-1.4em}
\label{table:3drec}
\end{table}

%% file: table_3.tex
\begin{table}[t!]
\vspace{0.0mm}
\setlength{\tabcolsep}{4pt} % general space between cols (6pt standard)
\centering
\resizebox{0.47\textwidth}{!}{%
%\sisetup{detect-weight=true}
\begin{tabular}{lccc|ccc}
\toprule
& \multicolumn{3}{c|}{C-Flow} & \multicolumn{3}{c}{C-Flow + cycle} \\
Method &  BPD$\downarrow$  & SSIM$\uparrow$  & IS$\uparrow$ &  BPD$\downarrow$  & SSIM$\uparrow$ & IS$\uparrow$  \\
\cmidrule(lr){1-1}  \cmidrule(lr){2-4} \cmidrule(lr){5-7} 
    segmentation $\rightarrow$ street views &  3.21 & 0.37 & 1.80 &  {\bf 3.17} &  {\bf 0.42} & {\bf 1.94}  \\
    segmentation $\leftarrow$ street views &   3.25 & 0.33 & 2.19 &  {\bf 3.05} & {\bf 0.36} & {\bf 2.23} \\
    structure $\rightarrow$ facades &  3.55 & 0.24 & {\bf 1.92} &   {\bf 3.54} & {\bf 0.26} & 1.69 \\
    structure $\leftarrow$ facades &  3.55 & {\bf 0.31} & {\bf 2.05} &  3.55 & 0.30 & 2.01 \\
    map $\rightarrow$ aerial photo & 3.65 & {\bf 0.19} & 1.52 &   3.65 & 0.17 & {\bf 1.62} \\
    map $\leftarrow$ aerial photo & 3.65 & 0.54 & 1.95 &   3.65 & {\bf 0.57} & {\bf 1.97} \\
    edges $\rightarrow$ shoes  &  1.70 & 0.66 & 2.40 &  {\bf 1.68} & {\bf 0.67} &  {\bf 2.43} \\
    edges $\leftarrow$ shoes  &    1.65 & 0.64 & 1.61 &  1.65 &  {\bf 0.65} &  {\bf 1.69} \\
\bottomrule
\end{tabular}
}
\caption{\small{{\bf Conditional image-to-image generation.} Evaluation of C-Flow (plain) and C-Flow + cycle consistency loss in image-to-image mapping.}} 
\label{table:im2im}
\vspace{-1.4em}
\end{table}

%% file: 05_conclusions.tex
\vspace{-2mm}
\section{Conclusions}
\vspace{-2mm}
We have proposed C-Flow, a novel conditioning scheme for normalizing flows.
This conditioning, in conjunction with a new strategy to model unordered 3D point clouds, has made it possible to  address  3D reconstruction and rendering images from point clouds, problems which so far, could not be tackled with normalizing flows. Furthermore, we demonstrate C-Flow to be a general-purpose model, being also applicable to many more multi-modality problems, such as  image-to-image translation, style transfer and image content edition. To the best of our knowledge, no previous model has   demonstrated such an adaptability.

\vspace{-2mm}
\section*{Acknowledgements}
\vspace{-2mm}
This project was done during an internship at Google. It is also partially supported by the EU project TERRINET: The European robotics research infrastructure network H2020-INFRAIA-2017-1-730994.